%% file: counterfactuals_reconsidered_ver2_non_blind.tex
\documentclass{article}
\usepackage{natbib}

\usepackage{xcolor} 
\usepackage{graphicx}
\usepackage{hyperref}
\usepackage{listings}
\usepackage{thm-restate}
\usepackage{amsmath}
\usepackage{amsfonts}
\usepackage{MnSymbol}

\usepackage{mathtools}

\input{defn}

\newtheorem{definition}{Definition}
\newtheorem{example}{Example}
\newtheorem{theorem}{Theorem}
\newtheorem{corollary}{Corollary}
\newtheorem{proposition}{Proposition}

\begin{document}

\title{Causal Counterfactuals Reconsidered}
\author{Sander Beckers\thanks{Preprint: currently under review}\\University College London\\srekcebrednas@gmail.com}
\date{}

\maketitle

 \begin{abstract}
{I develop a novel semantics for probabilities of counterfactuals that generalizes the standard Pearlian semantics: it applies to probabilistic causal models that cannot be extended into realistic structural causal models and are therefore beyond the scope of Pearl's semantics. This generalization is needed because, as I show, such probabilistic causal models arise even in simple settings. My semantics offer a natural compromize in the long-standing debate between Pearl and Dawid over counterfactuals: I agree with Dawid that universal causal determinism and unrealistic variables should be rejected, but I agree with Pearl that a general semantics of counterfactuals is nonetheless possible. I restrict attention to causal models that satisfy the Markov condition, only contain realistic variables, and are causally complete. Although I formulate my proposal using structural causal models, as does Pearl, I refrain from using so-called response variables. Moreover, I prove that my semantics is equivalent to two other recent proposals that do not involve structural causal models, and that it is in line with various comments on stochastic counterfactuals that have appeared in the literature more broadly. Throughout I also reflect on the universality of the Markov condition and explore a novel generalization of causal abstractions.}
\end{abstract}

\section{Introduction}\label{sec:int}

The Pearlian semantics for causal counterfactuals has dominated the field of causality for the last quarter of a century \citep{galles98,halpern00,pearl:book2}. This semantics owes part of its success to the fact that Pearl's causal modelling framework is rich enough to even express a formal semantics of counterfactuals: it merges a logical framework with a statistical one and extends both of these with a notion of an intervention that naturally fits conditional stipulations of the sort that counterfactuals consist of. Other -- somewhat rival, mostly compatible -- causal modelling frameworks exist, but none of them are equally expressive and none have put forward an alternative semantics of counterfactuals that is equally developed. 

Concretely, the graphical framework of  \cite{spirtes00} simply does not concern itself with causal counterfactuals at all. The Potential Outcomes framework \citep{rubin74,holland86,hernan20}, to the contrary, is built on causal counterfactuals, but it does not offer a fully general semantics, let alone a formal one. Furthermore, what it has to say about counterfactuals is almost entirely in agreement with Pearl's framework. Lastly, and most importantly, the decision-theoretic framework of \cite{dawid00} offers no alternative semantics of counterfactuals because it claims both that no such semantics is required for causal inference and that the Pearlian semantics is grounded in metaphysical assumptions about the world that are empirically unwarranted. Dawid's criticism has survived the last quarter of a century alongside the popularity of the Pearlian semantics, and the debate between the two sides has recently re-emerged from its state of relative dormancy \citep{mueller23,dawid23}. 

I believe a compromise in the sensible middle between their respective points of view is long overdue, and the aim of this paper is to offer one. I do so by adopting and expanding on Dawid's criticism, but instead of letting this result in a rejection of a semantics of counterfactuals altogether, I use it as inspiration to build up an alternative semantics. At the heart of the disagreement between Dawid and Pearl lies Pearl's view that any probabilistic causal model over a set of variables can be meaningfully transformed into a deterministic causal model by adding further, unobserved, variables. 
This transformation can then be exploited by reducing the semantics of counterfactuals in probabilistic models to the semantics in their deterministic counterparts. The entire debate therefore hinges on the status of these additional so-called {\em response variables} that Pearl relies on to obtain a deterministic model. In some cases these variables can be given a straightforward realistic interpretation, and if this occurs the Pearlian semantics is uncontroversial. Pearl assumes that they can be given a realistic interpretation in {\em all} cases, and thus his semantics is justified as a general semantics for all causal models. Dawid, on the other hand, denies this. He claims that in many cases no such realistic interpretation exists, and he makes the further claim that in all such cases counterfactuals are meaningless. 

The first main result of this paper is to show that Dawid's first claim is true, even when judged by the standards that Pearl sets himself. The second result is the development of a semantics for causal counterfactuals in those cases where no realistic extension of a probabilistic causal model exists. Concretely, I explicitly restrict attention to causal models that only contain realistic variables and that are causally complete, and yet need not be deterministic. Such causal models are not subject to Dawid's criticism, but fall outside of the scope of the Pearlian semantics, and therefore the challenge is to develop and defend a semantics that applies to them. 


This is not the first alternative semantics of this kind. Both \cite{galhotra24} and myself \citep{beckers25} have recently -- and concurrently -- developed a semantics for probabilities of counterfactuals that also applies directly to probabilistic causal models. I prove that all three semantics are equivalent. The benefit of the semantics in its current formulation and justification is that it shows both how exactly it results from the Pearlian semantics once we strip it off its unwarranted assumptions, and how it generalizes ideas that have shown up in the potential outcomes literature throughout the years. As will become clear, the crux of this idea is that if we take seriously the completeness of a causal model, then so-called {\em potential outcome variables} have to be mutually independent. Crucially, this idea has actually been suggested by Dawid himself on various occasions, further illustrating that my semantics offers a natural compromise between both sides of the debate. 

The paper proceeds as follows. I start out in Section \ref{sec:real} by laying out what it means for causal models to consist entirely of realistic variables and what it means for such a set of variables to be complete. In Section \ref{sec:cm} I define probabilistic causal models, adopting the causal Markov condition per assumption. I do so because it is required for my semantics, but -- as I later explain -- I am sympathetic to those who have criticized its universal applicability. I present the Pearlian semantics for deterministic causal models in Section \ref{sec:pearl}, and show in Section \ref{sec:nondet} that even in simple cases nondeterministic causal models emerge that cannot be extended into a realistic deterministic model. Given that this result undermines the universal applicability of the Pearlian semantics, I take it to be first main contribution of this paper. I briefly explore in Section \ref{sec:abs} how this result could be used to generalize the nascent field of causal abstractions, but set this exploration aside for future work. Section \ref{sec:pcnon} adds the final puzzle to the Pearlian semantics, namely the reduction of probabilistic models to a very specific type of deterministic models, so-called canonical models. My own semantics is developed in Section \ref{sec:poscm} by taking the Pearlian approach, but ridding it entirely from all unrealistic components in a manner that takes inspiration from the potential outcome framework. The equivalence of my semantics to my previous proposal and to another recent approach is then established in Section \ref{sec:equiv}, as well as its relation to various occasions where similar ideas have shown up in the literature.

\section{Realistic Causally Complete Causal Models}\label{sec:real}

\dfn
A signature $\cal S$ is a tuple $({\bf U},{\bf V},\R)$, where ${\bf U}$
is a set of \emph{exogenous} variables, ${\bf V}$ is a set 
of \emph{endogenous} variables,
and $\R$ a function that associates with every variable $Y \in  
{\bf U} \union {\bf V}$ a nonempty set $\R(Y)$ of possible values for $Y$
(i.e., the set of values over which $Y$ {\em ranges}).
If ${\bf X} = (X_1, \ldots, X_n)$, $\R({\bf X})$ denotes the
Cartesian product $\R(X_1) \times \cdots \times \R(X_n)$.
\edfn

As a causal model will be a model of the causal relations between the variables occuring in a signature, I impose two crucial informal conditions on a signature for it to fulfill this role. I call signatures that satisfy both conditions {\em realistic}, and similarly I call a causal model {\em realistic} if its signature is. Unless mentioned otherwise, I assume throughout that all signatures are realistic. 

First, a condition that is common to all causal modelling approaches, is that ${\bf V}$ should consist of variables that are -- either directly or indirectly -- observable. This means that each $Y \in {\bf V}$ is to be interpreted as representing a {\em real} empirical property, as opposed to a purely formal property that is added to a causal model for mathematical convenience. I extend this latter condition also to the exogenous variables: although perhaps unobservable, they should at least correspond to some property of the world whose meaning we understand and we know to exist. This is motivated by a principle of conceptual hygiene that is implicit throughout much of scientific modelling, whereby one includes into a model only those elements that correspond to realistic features of the world. This principle is also endorsed by \cite{dawid00}, and \cite{galhotra24} make comments along similar lines.\footnote{Dawid states, for example, that setting up a model requires understanding ``what real world quantities are represented by variables appearing in the model'' \citep[p. 407]{dawid00}. Galhotra and  Halpern motivate their approach by considering situations where also the exogenous variables are observable.} As I discuss later on, Pearl's relation to this principle, on the other hand, is ambiguous, adopting it throughout most of his work but not everywhere.

Second, for any  ${\bf X} \subseteq {\bf V}$, ${\bf x} \in \R({\bf X})$ it should be possible, at least in principle, to conceive of an external intervention -- written as $do({\bf X} = {\bf x})$ -- on the system that sets the variables ${\bf X}$ to the values ${\bf x}$. As such the idea of interventions appears in various forms across other causal modelling approaches as well. The Pearlian approach stands out by assuming that {\em all} interventions (on endogenous variables) are well-defined, and I here adopt this condition as well. This has major implications for the interpretation of any two variables $X,Y \in {\bf V}$: it means that the realistic properties expressed by each of them are entirely separable, meaning that we can give empirical meaning to any combination $X=x, Y=y$ of their values. For one thing, this rules out the existence of any logical relation between the meaning of $X$ and $Y$. The same implication follows for the exogenous variables from the fact that any setting ${\bf U}={\bf u}$ is taken to be well-defined. Furthermore, since any intervention $do({\bf X} = {\bf x})$ can occur in any context ${\bf U}={\bf u}$, it also follows that any $X \in {\bf V}$, $U \in {\bf U}$ are separable in this manner.

To be clear, one may adopt the empirical separability of variables as a condition without adopting the stronger condition that all interventions are well-defined. Nonetheless, my goal is not to argue that all causal models should be realistic, but merely to develop a semantics for probabilities of counterfactuals that is appropriate when causal models are realistic. For now I leave open the question whether the semantics can be meaningfully applied also to a broader class of causal models.

A realistic signature does not by itself guarantee that the causal relations over it can be expressed in a compact and fruitful manner. Specifically, there may exist confounding factors between two or more endogenous variables that are not captured by the exogenous variables present in the signature, and such confounding prevents the joint probability distribution to factorize in an elegant manner (i.e., it prevents the well-known Markov factorization). Therefore it is useful to also consider extending a signature with further exogenous variables. 

\dfn ${\cal S}_2=({\bf U}_2,{\bf V}_2,\R_2)$ is an {\em extension} of ${\cal S}_1=({\bf U}_1,{\bf V}_1,\R_1)$ if ${\bf V}_1={\bf V}_2$, ${\bf U}_1 \subseteq {\bf U}_2$, and $\R_1(Y)=\R_2(Y)$ for all $Y \in {\bf V_1} \cup ({\bf U}_1 \cap {\bf U}_2)$. 
\edfn

Lastly, as is standard, the causal relations between variables can be structurally represented as a directed acyclic graph over a signature.

\dfn A directed acyclic graph (DAG) $\G$ is {\em compatible} with a signature $\cal S$ if $\G$ contains one root node for each variable in ${\bf U}$ and one non-root node for each variable in ${\bf V}$. 
\edfn

\subsection{Causal Models}\label{sec:cm}

My definition of a causal model is identical to what   \cite{pearl:book2} and others call a {\em Causal Bayesian Network}, except that I also explicitly represent the exogenous variables in the graph, as do \cite{galhotra24}. This is again motivated by the aim to offer a semantics for complete and realistic models, for such models should not rely on any latent, and thus possibly unrealistic, variables. 

\dfn\label{def:PCM}
A \emph{causal model} $M$ is a triple $(\cal S,\G, P)$, 
where $\cal S$ is a signature, $\G$ is a DAG compatible with $\cal S$, and $P$ is a joint probability distribution over $\R({\bf U} \times {\bf V})$ that is Markov relative to $\G$, i.e., $P({\bf U},{\bf V})=\prod_{\{Y \in {\bf V}\}} P(Y | Pa_Y) \prod_{\{U \in {\bf U}\}} P(U)$.
Furthermore, for each ${\bf X} \subseteq {\bf V}$, ${\bf x} \in \R({\bf X})$, the joint distribution $P_{do({\bf X} = {\bf x})}$ is Markov relative to $\G_{\overline{\bf X}}$ and such that $P_{do({\bf X} = {\bf x})}({\bf X}={\bf x})=1$, i.e., it is given as
\begin{equation}\label{eq:int}
 P_{do({\bf X} = {\bf x})}({\bf U}={\bf u},{\bf V}={\bf v})=
  \begin{cases}
  0 \text { if }  {\bf x} \not \subseteq {\bf v}\\
    \prod_{\{Y \in {\bf V} \setminus {\bf X} \}} P(Y=y | {\bf Pa_Y}={\bf pa_Y}) \prod_{\{U \in {\bf U}\}} P(U=u) \text{ otherwise. }\end{cases}
\end{equation}
Here $(y,{\bf pa_Y}) \subseteq {\bf v}$ and $u \in {\bf u}$. As usual, $\G_{\overline{\bf X}}$ is constructed out of $\G$ by removing arrows coming into ${\bf X}$.
\edfn

Although I have built in the Markov condition -- that $P$ is Markov relative to $\G$ -- into the definition of a causal model, I do not assume that a realistic signature can always be extended into a realistic signature such that the Markov condition holds. Pearl on the other hand does adopt this assumption, and it forms a fundamental part of his framework.\footnote{Concretely, he states that ``Once we acknowledge the existence of latent variables and represent their existence explicitly as nodes in a graph, the Markov property is restored.'' \citep[p. 44]{pearl:book2}.} This assumption has been criticized both on philosophical \citep{cartwright99} and statistical grounds \citep{dawid10}, and I believe rightly so. We return to this topic in Section \ref{sec:nondet}, where I show that the assumption is false, even in a deterministic world. However, as with Pearl, the Markov condition plays a fundamental role in my semantics, and it is not at all clear to me how to generalize it beyond this setting. Therefore I here simply assume that we are dealing with a signature for which a correct causal model in the form of Definition \ref{def:PCM} exists. 

A necessary requirement for the Markov condition to hold, is that a signature contains sufficient variables to {\em screen off} the variables in the signature from all common causes outside of the signature.\footnote{This requirement is necessary only if one is considering {\em all} probability distributions compatible with the signature and graph, because some specific distributions might violate the {\em faithfulness} condition, according to which the only independencies are those implied by the Markov condition. When faithfulness is violated, the Markov  condition may hold despite the existence of a common cause outside of the signature that is not blocked by the variables in the signature.} This is why such a signature is said to be {\em causally sufficient} \citep{spirtes00}.  (Whether this requirement is also sufficient in the non-causal sense, meaning that it suffices to establish the Markov condition, is precisely the subject of the aforementioned debate.) Such sufficiency is concerned only with common causes, but remains silent on factors outside of the model that are causes only of a single endogenous variable. Since such variables could contain information relevant to counterfactuals, any model that excludes them can offer only a partial semantics of probabilities of counterfactuals at best. Therefore I further restrict attention to models with signatures that are {\em causally complete}, meaning that the exogenous variables screen off the endogenous variables from all of their remaining causes.

For every realistic signature there exists some realistic extension such that it is causally complete. (One could, for example, simply add a single exogenous variable that represents the aggregation of all missing causes.)  Of course in practice it may be very hard to find a causally complete signature, and therefore the practical utility of my semantics is limited to those cases where we can in effect establish causal completeness. One important class of such cases is when the causal model describes a system that we have constructed ourselves. As pointed out also by  \cite{galhotra24}, this includes both all computer programs (including AI systems) as well as manufacturing processes of various kinds. For example, elsewhere \cite{beckers25b} I offer an analysis of Large Language Models by interpreting them as causally complete causal models and then applying my semantics from \cite{beckers25}, which I later on show to be equivalent to the one I develop here.

I now present Pearl's semantics for probabilities of counterfactuals, which is by far the most popular one. In first instance we restrict to the case of deterministic models, as the nondeterministic case builds on the deterministic one.

\section{Pearlian Counterfactuals for Deterministic Models}\label{sec:pearl}

 \cite{pearl:book2} assumes that the world is causally deterministic, or at least, he restricts the application of his approach to domains that are  causally deterministic. Interestingly, one important reason he offers for doing so is precisely because he believes that it is a necessary prerequisite for defining counterfactuals.\footnote{Concretely, he states that ``In this book, we shall express preference toward Laplace’s quasi-deterministic conception of causality and will use it, often contrasted with the stochastic conception, to define and analyze most of the causal entities that we study.'' Further on he states that ``certain concepts ... can be defined only in the Laplacian framework. ... These so-called {\em counterfactual} concepts will require a synthesis of the deterministic and probabilistic components embodied in the Laplacian model. '' \citep[p.26-27]{pearl:book2}} My aim is to show him wrong. 

A domain is understood to be {\em causally deterministic} whenever there exists a causally complete realistic signature whose variables represent the properties of the domain, and a causal model over this signature such that all the probabilities of endogenous variables conditional on their parents are extremal, i.e., only take value in $\{0,1\}$. For such a model the only non-extremal probabilities are to be found in the distribution of $P({\bf U})$, and therefore the model takes on the form of a {\em structural causal model} (Pearl simply uses the term ``causal model'', see Def. 2.2.2 in \citep{pearl:book2}).

\dfn\label{def:scm}
A \emph{structural causal model} -- SCM -- $M$ is a 4-tuple $(\cal S,\G,\F, P_{\bf U})$, 
where $\cal S$ is a signature,  $\G$ is a DAG compatible with $\cal S$, and
$\F$ defines a  function that associates with each endogenous
variable $X$ a \emph{structural equation} $F_X$ taking on the form $X = f_X(\bf{Pa_X})$, where $\bf{Pa_X} \subseteq ({\bf U} \cup {\bf V} - \{X\})$ are the {\em parents} of $X$ as they appear in $\G$, and $f_X$ is a function $f_X: \R(\bf{Pa_X}) \rightarrow \R(X)$. Lastly, $P_{\bf U}$ is a joint probability distribution over $\R({\bf U})$ such that all $U, U' \in {\bf U}$ are mutually independent.\edfn

Pearl's motivation for defining an SCM in this form is entirely in line with my focus on realistic causally complete models. First, the reason that the exogenous variables need to be mutually independent is that this guarantees the signature to be causally sufficient, from which the Markov condition follows. Second, since the value of each endogenous variable $V$ is uniquely determined by $f_V$, there are no missing parents of $V$ at all. Third, we can find comments throughout Pearl's work that support the realistic nature of exogenous variables. For example,  \cite{tian00} 
 state [p. 309]:
 
 \begin{quote} The stochastic nature of the data is viewed as emerging from our ignorance of the detailed experimental conditions that prevailed in the study. The exogenous variables in ${\bf U}$ represent these missing details, and include the physiology and previous history of each person, his/her mental and spiritual attitude, as well as the time and manner in which the exposure occurred. In short, ${\bf U}$ summarizes all the factors which ``determine'' in the classical physical sense the outcome of the study. $P({\bf u})$ summarizes our ignorance of those factors.\end{quote}


I fully agree with Pearl on the semantics of probabilities of counterfactuals in realistic SCMs. I here present these semantics in a somewhat unorthodox fashion in order to ease comparison with other semantics later on, but my presentation is easily seen to be equivalent to Pearl's standard semantics (and the subsequent cleaned up version presented by \cite{bareinboim22}). For starters, we define a semantics for probabilities of basic counterfactuals.

\dfn\label{def:pcp} Given an SCM $M$ and some ${\bf u} \in \R({\bf U})$, ${\bf v} \in \R({\bf V})$ such that $P_M( {\bf U}={\bf u},{\bf V}={\bf v}) >0$, for any ${\bf X} \subseteq {\bf V}$, ${\bf x} \in \R({\bf X})$, we define the {\em probability of a basic counterfactual} as the following distribution over ${\bf V}$:
\begin{equation*}
 P_M({\bf V}_{{\bf X} \gets {\bf x}} | {\bf U}={\bf u}, {\bf V}={\bf v})=P_{do({\bf X} = {\bf x})}({\bf V} | {\bf U}={\bf u})
\end{equation*}
We write $P_M({\bf y}_{{\bf x}} | {\bf u},{\bf v})$ as shorthand for $P_M(({\bf Y}={\bf y})_{{\bf X} \gets {\bf x}} | {\bf U}={\bf u}, {\bf V}={\bf v})$ for ${\bf Y},{\bf X} \subseteq {\bf V}$.
\edfn

Importantly, given the nature of an SCM, all such probabilities are extremal ($1$ or $0$), and are computed as follows: 
\begin{itemize}
\item remove the equations for ${\bf X}$ from $\F$ and replace them with ${\bf X}={\bf x}$,
\item solve the novel equations following the partial order given by $\G$ for the setting ${\bf U}={\bf u}$,
\item return whether this unique solution is equal to ${\bf V}$ or not.
\end{itemize}

What makes these expressions counterfactual, is that the intervention $do({\bf X}={\bf x})$ may conflict with the original solution $({\bf U}={\bf u}, {\bf V}={\bf v})$ to the model, and therefore the resulting world $({\bf U}={\bf u},{\bf V}={\bf v}^*)$ is counterfactual relative to the original, actual, world. Probabilities of complex counterfactuals are now defined by taking probabilities of basic counterfactuals to be independent of each other conditional on the actual world, and then applying the law of total probability.\footnote{See \citep{beckers25} for a proof that this definition is equivalent to that of Pearl.}

\dfn\label{def:pc} Given an SCM $M$, we define {\em probabilities of complex counterfactuals} as the family of distributions over counterfactual events of the form ${\bf y}_{{\bf x}}, \ldots, {\bf z}_{{\bf w}}$ for any ${\bf Y},{\bf X}, \ldots, {\bf Z},{\bf W} \subseteq {\bf V}$, as follows:
\begin{align*}
P_M({\bf y}_{{\bf x}}, \ldots, {\bf z}_{{\bf w}})=\sum_{\{{\bf u},{\bf v}\}} P_M({\bf y}_{{\bf x}} | {\bf u},{\bf v}) \ldots P_M({\bf z}_{{\bf w}} | {\bf u}, {\bf v}) P({\bf u}, {\bf v}).
\end{align*}\edfn

\section{The Emergence of Nondeterminism}\label{sec:nondet}

My disagreement with Pearl arises once we consider realistic and causally complete causal models that are {\em not} SCMs. Pearl simply assumes that such models do not exist, because of the following reason. Given any realistic signature and any causal model over it, if the model is not an SCM, then he assumes that we can extend its signature so that the corresponding extended model is. This assumption underlies all of the many results regarding the partial identifiability of probabilities of counterfactuals in the literature (see for example \citep{balke94,tian00,zhang22}). The problem is that in many cases there does {\em not} exist an extension such that the additional variables are realistic. 

To show this, we need to assume that in some domains there exists a fundamental signature describing it, by which I mean a signature made up of variables that are maximally fine-grained. In such a signature, each value of each variable represents a property that cannot be realistically subdivided into further distinct properties.\footnote{Although Pearl does not mention the idea of fundamental variables, within the context of his defense of the Markov condition he does allude to there being a most fine-grained level, stating: ``We can start in the deterministic extreme, where all variables are explicated in microscopic detail and where the Markov condition certainly holds.'' \citep[p. 44]{pearl:book2}.} As the following example shows, such cases do not always allow for realistic extensions into an SCM.

\xam\label{ex:nondet} Say we have a fundamental realistic SCM $M_1$ with signature ${\cal S}_1=({\bf U}_1={\bf U},{\bf V}_1=\{Y\},\R_1)$, a single equation $Y=f_Y({\bf U})$, and a distribution $P({\bf U})$, such that $P({\bf U}={\bf u}) > 0$ for all ${\bf u} \in \R_1({\bf U})$ and $P(Y=y) > 0$ for all $y \in \R_1(Y)$. Furthermore, assume that $|\R_1(Y)| > 2$. (Note that this implies also $|\R_1({\bf U})| > 2$.) 

Consider any ${\bf u}' \in \R_1({\bf U})$, and define the binary variable $Z$ that corresponds to the property whether ${\bf U} = {\bf u}'$ (in which case $Z=1$) or not (in which case $Z=0$). Given that ${\bf U}$ is realistic, so is $Z$. Also, given that ${\bf U}$ and $Y$ are separable, so are $Z$ and $Y$. Therefore ${\cal S}_2=({\bf U}_2=\{Z\},{\bf V}_2=\{Y\},\R_2(Z)=\{0,1\},\R_2(Y)=\R_1(Y))$ forms a realistic signature. 

The correct causal model $M_2$ over ${\cal S}_2$ is not an SCM, for the probabilities of the child conditional on all of its parents are not restricted to $\{0,1\}$. Concretely, let $y_1,y_2,y_3$ be distinct members of $\R_1(Y)$ with $y_1=f_Y({\bf u}')$, and let $P(Y=y_2 | Z=0) =p$. Given that $P(Y=y) > 0$ for all $y \in \R_1(Y)$, it holds that $0 < P(Y=y_2) < 1$. Furthermore, $P(Y=y_2) = P(Y=y_2 | Z=0)P(Z=0)+ P(Y=y_2 | Z=1)P(Z=1) = p P(Z=0)+  0 P(Z=1)=p P(Z=0)=p P({\bf U}\neq {\bf u}')$. So $0<p P({\bf U}\neq {\bf u}')<1$. Given that $0 <P({\bf U}={\bf u}') < 1$, also $0< P({\bf U}\neq {\bf u}') < 1$, and thus $0 < p <1$.

We now prove by a reductio that there does not exist a realistic causally complete extension of $M_2$ such that it is an SCM. Assume that ${\cal S}_3=({\bf U}_3=\{Z\} \cup {\bf W},{\bf V}_3=\{Y\},\R_3)$ forms a realistic signature and that the correct causal model $M_3$ over it is an SCM. Thus, we have that there exists some surjective function $g_Y: \R(Z) \times \R({\bf W}) \rightarrow \R_1(Y)$ such that there is a structural equation $Y=g_Y(Z,{\bf W})$. Since $| R(Z)|=2$ and $|\R(Y)| > 2$, it must be that $|\R(W)| \geq 2$.

Given that $Z=1$ -- per definition -- corresponds to ${\bf U} = {\bf u}'$, and given that $y_1=f_Y({\bf u}')$, we know that $g_Y(1,{\bf w})=y_1$ for all ${\bf w} \in \R({\bf W})$. This means that the single value  ${\bf U} = {\bf u}'$ corresponds to multiple values $(1,{\bf w}_1), \ldots, (1,{\bf w}_n)$. Since ${\bf W}$ is realistic, each ${\bf w} \in \R({\bf W})$ expresses a real and distinct property of the domain.\footnote{That different values of a variable should be distinct is in fact an accepted condition on causal variable choice that goes beyond realistic variables, see \citep{halpernhitchcock11}.} Therefore we get that the property expressed by ${\bf U} = {\bf u}'$ corresponds to several real and distinct properties, which contradicts that ${\bf U}$ is a set of fundamental variables.
\exam

What this example shows, is that for realistic models, the existence of a deterministic model at the fundamental level does not in general imply the existence of a deterministic model at a more abstract level. Specifically, whether or not a deterministic model exists at a non-fundamental level depends on how one chooses the abstraction $\tau$ mapping the values of lower-level variables to those of higher-level variables, and the study of this choice has in recent years become an established part of the causal landscape \citep{rubenstein17,beckers19,beckers20a,zennaro23a}. Throughout most of the work on {\em causal abstraction} (as it is called), however, the restriction to realistic models is either left implicit or is simply entirely absent, and therefore the above negative result had not yet been established. Given its implications for the study of counterfactuals, as well as for broader philosophical issues regarding the emergence of nondeterminism more generally, I consider this insight to be the first main contribution of this paper. 

As we arrived at this conclusion by relying on there being a fundamental description of a domain, in principle one could resist it by offering an argument that there never exists a fundamental level of description in any domain that is subject to causal relations. I doubt a persuasive argument of this kind could be given, but even if it could, it would seem unnecessarily restrictive to let our entire semantics of causal counterfactuals rely on this rather unorthodox view. 

The above example can easily be modified to show something stronger, namely that Pearl's claim about the universality of the Markov condition also fails to hold. Concretely, the following trivial example shows that there does not always exists a realistic and complete causal model satisfying the Markov condition, for much the same reason as there does not always exist an SCM. 

\xam\label{ex:markov}
Imagine a very simple instance of a fundamental and realistic SCM $M_1$ from Example \ref{ex:nondet}, where we also add a second effect variable $A$.  Concretely, the signature is ${\cal S}_1=({\bf U}=\{X\},{\bf V}=\{Y,A\},\R_1)$, where all variables range over $\{1,\ldots, 100\}$, with the equations $Y=X$ and $A=X$, and uniform $P(X)$. So $Y$ and $A$ share the common parent $X$, and are mutually independent conditional on $X$, in line with the Markov condition. Note also that $P(A=Y) =1$. 

Now consider the binary variable $Z$, which expresses the property corresponding to whether $X \leq 40$ (in which case $Z=0$) or not (in which case $Z=1$), and consider the realistic signature ${\cal S}_4=({\bf U}=\{Z\},{\bf V}=\{Y,A\},\R_4(Z)=\{0,1\},\R_4(Y)=\R_1(Y),\R_4(A)=\R_1(A))$ and corresponding causal model $M_4$. 

$Z$ is a common parent of $Y$ and $A$ in $M_4$. (This follows from the fact that there is a non-zero direct effect of $Z$ on $Y$, and on $A$. Concretely, for all $a \in \R(A)$, $E(Y | Z=1, do(A=a)) - E(Y | Z=0, do(A=a)) \neq 0$, and similarly for $Y$ and $A$ reversed.) Furthermore, $Z$ is the only parent of $Y$, and the only parent of $A$, as neither are parents of each other.

But $P(Y | Z,A) \neq P(Y | Z)$, and thus the Markov condition does not hold. (For example, $P(Y \neq 10 | Z=0, A=10) = 0$, whereas $P(Y \neq 10 | Z=0) = 39/40$.) Given that $A$ perfectly predicts $Y$ (and vice versa), the only way to extend $M_4$ so that Markov is restored, is to extend it into an SCM. We can apply exactly the same  reasoning as we did in Example \ref{ex:nondet} to conclude that no such realistic extension is possible.
\exam

The obvious reply to both of these negative results is that one should make sure to choose one's variables wisely, so that these situations cannot arise. Indeed, that is the standard response given to \cite{cartwright99}'s alleged counterexamples to the Markov condition (at least, to those examples that do not rely on quantum phenomena) \citep{hausman99,spirtes00}.
The latter's discussion of Salmon's pool example offers a simple illustration of this. Imagine a cue ball colliding with two adjacent balls that are both lying at $45^{\circ}$ angles from the left and right bottom pockets, respectively. If we represent the collision with a binary variable, then learning that the left ball went into its pocket affects the probability that the right one went into the other pocket, even when conditioning on there being a collision, and thus the Markov condition is violated. But the Markov condition can be regained by representing the collision with a more fine-grained variable that specifies the precise momentum of the cue ball when it hits the balls. 
The problem is that in the absence of non-question begging criteria to determine what makes a choice of variables appropriate, this reply is of no practical use whatsoever. The fact remains that merely having found a realistic and complete signature does not allow us to conclude that the Markov condition holds, let alone that the causal model over this signature has to be deterministic. I suggest therefore that the debate about the universality of the Markov condition be replaced with the -- more useful -- search for such criteria. 

Given that an SCM trivially satisfies the Markov condition, all cases of causal abstraction of a realistic signature for which there is no Markov-satisfying realistic extension are also cases for which there is no extension into a realistic SCM. The semantics for causal models as I have defined them -- i.e., as satisfying the Markov condition -- will therefore be a generalization of Pearl's semantics for SCMs, but it leaves open the possiblity that one could generalize even further to also include semi-Markovian realistic causal models. (Semi-Markovian refers to acyclic causal models that need not satisfy the Markov condition.)

\subsection{Generalizing Causal Abstractions}\label{sec:abs}

Before getting back to Pearlian counterfactuals, I briefly consider a natural proposal that suggests itself for handling the kind of emergence of nondeterminism illustrated in Examples \ref{ex:nondet} and \ref{ex:markov}: since probabilities of counterfactuals and probabilities of interventions are well-defined within the realistic and fundamental SCM $M_1$, we could reduce such probabilities in the induced causal model $M_2$ to their counterparts in $M_1$. If effective, this proposal would generalize the definition of a causal abstraction beyond the case where both the lower level and the higher level have to be SCMs, whilst still offering semantics for all three layers of the causal hierarchy at the higher level. Given the importance of this field of research, I consider this a worthwhile avenue to pursue. However, I set it aside for a different occasion, for three reasons. 

First, such a proposal is useful only in case such a realistic SCM exists. This is warranted whenever the assumption of causal determinism applies, but there is little scientific evidence to suggest that it always does. 
In fact, according to one popular view among philosophers of science, the very concept of causation itself does not occur within fundamental physics and only applies at the macroscopic scale \citep{sep:causation}. To be clear, and setting quantum mechanics aside (as does Pearl), this view is entirely consistent with an assumption that the world is fundamentally deterministic. The point is that having a complete description in terms of an SCM is a significantly higher bar to pass than merely having some complete description. Furthermore, even if determinism is deemed plausible (outside the quantum domain), it would be unnecessarily restrictive to let our entire causal methodology rest on assuming that it is universally applicable.\footnote{As we will see below, much the same point is made by \cite{dawid00,dawid12}. Cartwright  similarly states that ``our evidence is not sufficient for universal determinism. To the contrary, for most cases of causality we know about, we do not know how to fit even a probabilistic model, let alone a deterministic one. The assumption of determinism is generally either a piece of metaphysics that should not be allowed to affect our scientific method, or an insufficiently warranted generalisation from certain kinds of physics and engineering models.'' \citep[p.13]{cartwright99}}

Second, even if the existence of some kind of realistic SCM is not in doubt, in many cases the details of any such candidate SCM and the variables appearing in it are too far removed from the variables appearing in a causal model of interest to connect the two by a stable mapping $\tau$. This occurs for example whenever the variables in our causal model are intentionally left vague or partially indeterminate, as is the case in much of the social sciences as well as with most concepts in natural language.\footnote{See \citep{edgington24} for a philosophical analysis of such concepts and how they result in indeterminate counterfactuals. Both the formal semantics that I develop and its motivation are very much in line with her views on the topic.} The absence of an abstraction mapping $\tau$ between such ambiguous concepts and some more fine-grained variables underlying it is not just a reflection of our ignorance, but also reflects the absence of any {\em decision} on our part as language users to construct and define such a mapping. There exists an inexhaustible supply of such concepts. Think of examples as diverse as ``civil war", ``inflation'', ``being grateful'', ``acting like an idiot'', ``game playing'', or any other socially constructed concept: all of these can fulfil causal roles, and for none of them have we decided how their meanings should relate to microscopic descriptions of the world in a manner that is sufficiently stable for there to be some $\tau$.\footnote{\cite{mosse25} have recently developed a proposal for interpreting social constructs as causal factors by invoking the idea of abstractions. Their discussion is mostly informal, however, and they rely on the assumption that we have at least partially decided what such a mapping looks like. This suggests that there is room for an approach where we combine having a partial abstraction with a remaining degree of vagueness.} Relatedly, in many cases the precise lower level realization of the variables in our model depend on highly contingent context-dependent features that are not part of the domain of interest, and therefore I prefer interpreting the probabilistic causal model at face value. For example, in principle the behavior of pool balls is subject to entirely deterministic processes, yet we hardly ever have use for a model that depends on making these processes explicit. I \cite{beckers25b} develop one such domain in detail by showing how the intended interpretation of probabilistic Large Language Models can be made formally precise using probabilistic causal models of the kind we here discuss. The basic idea is that although the implementation of any probabilistic model requires using deterministic pseudo-random number generators, these contingent deterministic processes are usually seen as mere implementation-level details that should be ignored. The same holds for probabilistic causal models across a variety of domains, and the above proposal would therefore not be applicable. 

Third, as I now explain, developing the proposal requires overcoming a significant obstacle that stands mostly orthogonal to the topic at hand. 

Roughly put, an abstraction generalizes the idea from the above examples by taking an SCM $M_1$, a surjective mapping $\tau: \R_1({\bf U}_1 \times {\bf V}_1) \rightarrow \R_2({\bf U}_2 \times {\bf V}_2)$, and then tries to construct an SCM $M_2$ over the signature ${\cal S}_2=({\bf U}_2,{\bf V}_2,\R_2)$ that is consistent with $M_1$. The most common type of abstraction mapping (and the only one that can satisfy our realistic conditions) is a {\em constructive abstraction} \citep{beckers19}, where $\tau$ can be decomposed so that for each $Y \in {\bf U}_2 \cup {\bf V}_2$ there is a mapping $\tau_{Y}: \R_1({\bf Y}_1) \rightarrow \R_2(Y)$ for some ${\bf Y}_1 \subseteq {\bf U}_1 \cup {\bf V}_1$. The proposal consists of generalizing this idea so that $M_2$ is allowed to be a causal model that is not necessarily an SCM, and to define probabilities of counterfactuals in $M_2$ by inverting $\tau$ and relying on the probabilities of counterfactuals in $M_1$. This means we need to add what might be called a family of {\em inversion distributions} $P^I({\bf Y}_1 | \tau_{Y}({\bf Y}_1))$ for each $Y$, representing the probability that any given high level value is realized as any of its compatible low level values. In addition, we need to define a family of distributions that map high level interventions to low level interventions for each $Y$: $P^I(do({\bf Y}_1) | do(\tau_{Y}({\bf Y}_1)))$.\footnote{I consider a similar type of distribution in \citep{beckers20a}, but only when restricted to the SCM case.} The latter means we need to offer a semantics for {\em disjunctive interventions}, for $P^I(do({\bf y}_1) | do(y))$ can be written as: 
$$P^I(do({\bf y}_1) | do(\bigvee_{\{{{\bf y}_1}' | \tau_Y({{\bf y}_1}')=y\}} {{\bf y}_1}')).$$

\cite{pearl17} has addressed the special instance of this problem where we have just a single causal model to deal with, only to conclude that any solution which aims to offer precise values of such probabilities is forced to rely on ad hoc, metaphysical, assumptions. The situation where two models are involved is significantly more complicated, and therefore I do not delve further into it here. Ironically, in this context Pearl states ``that caution need be exercised when metaphysical extensions are taken literally, without careful guidance of decision making considerations'' \citep[p.4]{pearl17}. As we shall now see, this sounds remarkably similar to the message that Dawid has with respect to Pearl's semantics of probabilities of counterfactuals.

\section{Pearlian Counterfactuals in Nondeterministic Models}\label{sec:pcnon}
 
If we abandon realism, then Pearl's contention that any signature and any causal model over it can be extended into an SCM does hold, and forms the basis of Pearl's method to give semantics of probabilities of counterfactuals beyond SCMs. This method consists in extending a model with so-called {\em response variables}, that serve to ``pull out'' all the probability from the causal relations and place it instead on these additional exogenous variables \citep{balke94}. Recently  \cite{zhang22} have studied Pearl's method in its full generality, including causal models that do not satisfy the Markov condition. Once we restrict ourselves to causal models that do, their definitions simplify significantly. It will be convenient to also include a definition of a canonical frame as a template for all the SCMs that extend a given causal model.

\dfn\label{def:canon}
Given a causal model $M=(\cal S,\G, P)$, we define the {\em canonical frame} $(\cal S^C, \G^C,\F^C)$ of $M$ as follows.
\begin{itemize}
\item ${\bf V}^C={\bf V}$
\item ${\bf U}^C={\bf U} \cup_{\{X \in {\bf V}\}} \{U_X\}$
\item for each $X \in {\bf V}$: $\R(U_X)=\{f_X: \R({\bf Pa_X}^{\G}) \rightarrow \R(X)\}$ (here ${\bf Pa_X}^{\G}$ indicates the parents relation of $\G$.) 
\item $\G^C = \G \cup_{\{X \in {\bf V}\}} \{U_X \rightarrow X\}$
\item for each $X \in {\bf V}$: $\F^C$ associates $X$ with the equation $X = f_X(\bf{Pa_X}^{\G})$, where $f_X \in \R(U_X)$. 
\end{itemize}

Given an SCM $M'$ over the canonical frame, we say that $M'$ is a {\em canonical SCM} for $M$ if $P_{{\bf U}^C}({\bf U}) = P_M({\bf U})$, and
for each $X \in {\bf V}$, $P_M(X | {\bf Pa_X})=\sum_{\{f_X \in \R(U_X) | X=f_X({\bf Pa_X})\}} P_{{\bf U}^C}(U_X=f_X)$. \edfn

It is easy to show that each canonical $M'$ agrees with $M$ on all of its observational and interventional distributions over the endogenous variables, i.e., all distributions of the form $P_{do({\bf X} = {\bf x})}({\bf V})$. (Note that ${\bf X} = \emptyset$ results in $P({\bf V})$.) Therefore each of them is said to {\em induce} $M$. Pearl assumes that for any causal model $M$, one of its canonical SCMs represents the ground-truth. Probabilities of counterfactuals in causal models are then reduced to their counterparts in the true underlying SCM. Since knowledge of the true SCM is usually not available, such probabilities are in general not identifiable, and the best we can do is to investigate conditions under which we can derive their bounds \citep{balke94,tian00,zhang22}. The upshot is that counterfactuals are always well-defined, but often not identifiable due to a lack of knowledge. Here is a simple example to illustrate this reduction.

\xam\label{ex:illus} Say we have a causal model $M$ with binary endogenous variables $X$ and $Y$, a graph $X \rightarrow Y$, some values $0 < p < q <1$ such that $P(Y=1 | X=1)=p$ and $P(Y=1 | X=0)=q$, and some distribution $P(X)$. (We leave the exogenous variables that determine $X$ implicit.) In this simple case we get a four-valued response variable $U_Y$ so that $f_Y$ is either the identity function, the function that reverses the value of $X$, or the constant function that always returns $0$/$1$. We can represent this by letting $\R(U_Y)=\{0,1,2,3\}$ denote each of these respective options, so that $Y$ is determined by $X$ and $U_Y$ as follows: $Y=X$ if $U_Y=0$, $Y=\lnot X$ if $U_Y=1$, $Y=0$ if $U_Y=2$, and $Y=1$ if $U_Y=3$. Observe that any choice of $P(U_Y)$ such that $p=P(U_Y=0)+P(U_Y=3)$ and $q=P(U_Y=1)+P(U_Y=3)$ satisfies the requirement that $P(Y=1 | X=1)=p$ and $P(Y=1 | X=0)=q$. 

Probabilities of counterfactuals over $M$ are now defined as the corresponding probabilities over the true canonical SCM $M'$. Say for example we take $M'$ to be such that $P(U_Y=3)=0$, $P(U_Y=0)=p$, and  $P(U_Y=1)=q$, then this gives:
\begin{multline*}P_M((Y=0)_{X \gets 0} | X=1, Y=1)\overset{def}{=}P_{M'}((Y=0)_{X \gets 0} | X=1, Y=1)\overset{(1)}{=}\\ 
\sum_{\{u_Y \in \R(U_Y)\}} P_{M'}((Y=0)_{X \gets 0} | U_Y=u_Y,X=1, Y=1) P_{M'}(U_Y=u_Y | X=1, Y=1)\overset{(2)}{=}\\
P_{M'}((Y=0)_{X \gets 0} | U_Y=0,X=1, Y=1) \overset{Def. \ref{def:pcp}}{=} P_{do(X=0)}(Y=0 | U_Y=0) \overset{Eq. (\ref{eq:int})}{=}\\
 P(Y=0 | U_Y=0, X=0)\overset{(3)}{=}1.
\end{multline*}
Here (1) is an application of the law of total probability, (2) follows from standard conditioning given that $P(U_Y=3)=0$, and (3) follows from the equations described above.

If instead we take $M'$ such that $P(U_Y=3)=p$, $P(U_Y=0)=0$, and  $P(U_Y=1)=q-p$, then the same computation results in $P_M((Y=0)_{X \gets 0} | X=1, Y=1)=0$. As a result, if we are entirely agnostic regarding the true canonical model, the probabilities of basic counterfactuals for $Y$ are entirely unbounded.
 \exam

As I indicated earlier, Pearl's attitude towards realism is ambiguous, and this ambiguity is particularly crucial when it comes to the response variables $U_X$, because it is their generality that undergirds Pearl's semantics. Due to their ability to express any possible function from the domain of the parents to that of a child, they are general enough to encode any relevant causal factors that have been left out of $M$'s signature. Once realism is abandoned, however, they can also encode purely formal, artificial, properties that have no empirically verifiable counterpart in the world. Pearl (and those who adopt his semantics) now faces an unappealing dilemma. If response variables are interpreted realistically, then he is overlooking the negative result (Ex. \ref{ex:nondet}) from the previous section that such variables do no always exist. If interpreted artificially, then he is attributing properties to the world that are in fact purely formal properties of a conventional mathematical representation. 

In \citep[p. 1]{pearl11} Pearl makes a very revealing comment in this regard, as it overlooks the fundamental difference between both interpretations, by stating that the value of a response variable ``may stand either for the identity of a unit (e.g., a person’s name) or, more functionally, for the set of unit-specific characteristics that are deemed relevant to the relation considered.'' As pointed out also by  \cite{dawid12}, both disjuncts contain very different interpretations. If a response variable is nothing but a label to pick out a specific instance of a setting that instantiates the causal model, then it is purely artificial, whereas if it describes unit-specific characteristics, then it is real. 

Similar to my embrace of realistic causal models, \cite{dawid00} wants to expunge artificial variables from causal models, and therefore the artificial interpretation is a non-starter to him. The realistic interpretation, on the other hand, 

\begin{quote}
constitutes a very strong assumption of {\em determinism}: that, once we have measured these attributes (for a given unit), we will be able to predict,
{\em without error}, exactly what value that unit’s exposure $X$ will take, in response to either input value for $Z$ [(here $Z$ is a binary variable assumed to be the only endogenous parent of $X$)]. Such Laplacian determinism is out of favour as a general scientific or philosophical principle, and it seems odd to make it the cornerstone of a general theory of causality -- especially since there exist non-deterministic alternatives, such as the decision-theoretic approach mentioned above. \citep[p. 2]{dawid12}
\end{quote}

I now proceed to offer an alternative nondeterministic alternative, but unlike that of Dawid and  Didelez it does not shy away from counterfactuals. In fact, I present what appear to be three alternative semantics, and then show that they are all equivalent. The first of these is a novel proposal, whereas the latter two semantics are recent additions to the literature whose relation was so far not clearly understood.

\section{The Semantics Inspired by Potential Outcomes}\label{sec:poscm}

I start with developing a novel semantics, one that takes the Pearlian one as a blueprint but then rids it of any component that cannot be justified by realism. Interestingly, the idea for doing so lies in taking an SCM and replacing Pearl's response variables with a type of variables that are the hallmark of the other major -- and somewhat rival -- approach to causal models, namely the Potential Outcomes framework (PO) \citep{rubin74,holland86,hernan20}. Concretely, a single response variable $R_X$ can be equivalently represented as a set of {\em potential outcome} variables, $X_{\bf pa_X}$, one for each ${\bf pa_X} \in \R({\bf Pa_X})$. (An important note on terminology: in the literature the term potential outcome variable is sometimes applied more liberally, namely even when ${\bf Pa_X}$ does not contain {\em all} parents of $X$. Here we do not do so.) A potential outcome variable $X_{\bf pa_X}$ does as its name suggests: it expresses the outcome for $X$ if the potential value ${\bf Pa_X}={\bf pa_X}$ were to be realized. So if this potential value is in fact the actual value, then we have $X=X_{\bf pa_X}$.\footnote{The assumption that this identity holds is called {\em consistency}, and forms a fundamental building block of both the Pearlian and the PO framework. Although it may seem like a tautology, it takes on more substance outside of the safe confines of an SCM. Indeed, within the PO framework this assumption has to be justified, as there may exist situations in which it is not as obvious as it seems  \citep{vanderweele13,hernan16}.  \cite{pearl:book2}, instead, claims that it is not an assumption of his framework but rather a consequence that can easily be proven to hold. The problem is that this holds only if we take for granted that any causal model can be extended into a SCM, which is precisely what I have rejected in Section \ref{sec:nondet}.} Thus, all it takes to move from the one representation to the other is to interpret $X=f_Y({\bf pa_X})$ as $X=X_{\bf pa_X}$ if ${\bf Pa_X}={\bf pa_X}$, and vice versa. Pearl is well aware of this and makes use of this equivalence within the same context mentioned above \citep{pearl11}. We can make this idea fully precise by defining an alternative extension of a causal model into an SCM using such PO variables.

\dfn\label{def:noncanon}
Given a causal model $M$, we define the {\em PO-SCM} $M^{N}$ for $M$ as follows.
\begin{itemize}
\item ${\bf V}^{N}={\bf V}$
\item ${\bf U}^{N}={\bf U} \cup_{\{X \in {\bf V}\}} (\cup_{\{{\bf pa_X} \in \R({\bf Pa_X})\}} \{X_{\bf pa_X}\})$
\item for each $X_{\bf pa_X} \in {\bf U}^{N}$: $\R(X_{\bf pa_X})=\R(X)$
\item $\G^{N} = \G \cup_{\{X_{\bf pa_X} \in {\bf U}^{N}\}} \{X_{\bf pa_X} \rightarrow X\}$
\item for each $X \in {\bf V}$: $\F^{N}$ associates $X$ with the equations $X = X_{\bf pa_X}$ if ${\bf Pa_X}={\bf pa_X}$, which I write as $X = X_{\bf Pa_X}$.
\end{itemize}
Lastly, $P_{{\bf U}^{N}}({\bf U}) = P_M({\bf U})$, and
for each $X_{\bf pa_X} \in {\bf U}^{N}$, $P_M(X | {\bf pa_X})=P_{{\bf U}^{N}}(X_{\bf pa_X})$. We write $P^{N}$ for $P_{M^{N}}$.
\edfn

Similar to the canonical case, it directly follows that $M^{N}$ agrees with $M$ on all of its observational and interventional distributions, and thus induces $M$. Contrary to the canonical case, the PO-SCM is unique, and thus probabilities of counterfactuals in $M$ can simply be identified with those in $M^{N}$. I call this the {\em $PO$ semantics of probabilities of counterfactuals}. We return to our previous Example \ref{ex:illus} to illustrate how this works.

\xam\label{ex:illus2} Take the same causal model $M$ as before (again ignoring the exogenous variables for $X$). In $M^N$ we have two potential outcome variables, $Y_{1}$ and $Y_{0}$, and the equation for $Y$ is simply $Y=Y_X$. Given that 
$P_M(Y=1 | X=1)=p$ and $P_M(Y=1 | X=0)=q$, we get that $P^N(Y_1=1)=p$ and $P^N(Y_0=1)=q$. 

Probabilities of counterfactuals over $M$ are now defined as the corresponding probabilities over $M^N$. So we get, for example: 
\begin{multline*}P_M((Y=0)_{X \gets 0} | X=1, Y=1)\overset{def}{=}P^N((Y=0)_{X \gets 0} | X=1, Y=1)=\\ 
\sum_{\{(y_1,y_0) \in \R(Y_1 \times Y_0)\}} P^N((Y=0)_{X \gets 0} |  Y_1=y_1, Y_0=y_0,X=1, Y=1) P^N(Y_1=y_1, Y_0=y_0 | X=1, Y=1)\overset{(1)}{=}\\
\sum_{\{y_0 \in \R(Y_0)\}}  P^N((Y=0)_{X \gets 0} |  Y_1=1, Y_0=y_0,X=1, Y=1) P^N(Y_0=y_0 | Y_1=1)\overset{(2)}{=}\\
\sum_{\{y_0 \in \R(Y_0)\}} P^N((Y=0)_{X \gets 0} |  Y_1=1, Y_0=y_0,X=1, Y=1) P^N(Y_0=y_0)\overset{Def. \ref{def:pcp}}{=}\\
\sum_{\{y_0 \in \R(Y_0)\}} P_{do(X=0)}(Y=0 |  Y_1=1, Y_0=y_0) P^N(Y_0=y_0)\overset{(3)}{=}
P^N(Y_0=0)=1-q
\end{multline*}
Here (1) follows from the fact that $Y=Y_X$, (2) follows from the independence of exogenous variables, and (3) follows again from $Y=Y_X$ together with Eq. \ref{eq:int}. Contrary to the Pearlian analysis in Example \ref{ex:illus}, here all probabilities of counterfactuals are point-identified.
 \exam

The unicity of the PO-SCM is a consequence of two factors. First, we have no choice but to take $P_{{\bf U}^{N}}(X_{\bf pa_X})$ as equal to $P_M(X | {\bf pa_X})$ to get the distributions of $M$ and $M^N$ to be identical. Second, since the potential outcome variables are exogenous, they are independent, and thus the joint distribution is fully determined by the previous factor. 
This independence, however, was justified by invoking the Markov condition in the context of causally complete models with realistic variables, and thus one might wonder whether it should also apply to potential outcome variables. Note that in any given world exactly one of the values ${\bf pa_X}$ is realized, and thus only one potential outcome variable has a realistic interpretation, namely $X_{\bf pa_X}$. Let us call this the {\em actual outcome variable}, to be contrasted with the {\em counterfactual outcome variables}. Before some $X=x$ is realized in a world with ${\bf Pa_X}={\bf pa_X}$, $P(X_{\bf pa_X})$ represents the realistic property that the prior distribution of $X$ is $P(X | {\bf pa_X})$. After $X=x$ is realized, $X_{\bf pa_X}=x$ represents the realistic property that the posterior distribution of $X$ is $P(X=x | {\bf pa_X})=1$. All the counterfactual outcomes variables, on the other hand, are nothing but artificial artefacts of our deterministic representation, and thus they are irrelevant. The independence assumption is itself an artificial artefact of our representation that serves as a formal tool to achieve this irrelevance. 

Importantly, I am not the first to propose this independence assumption. In fact, in the few cases where counterfactuals in nondeterministic causal models have been discussed by other authors, this very independence assumption shows up in some form or other. We return to these cases in Section \ref{sec:equiv}, first we compare the PO-SCM to Pearl's canonical SCMs. 

The PO-SCM is not a canonical SCM, because it does not contain the canonical frame $(\cal S^C, \G^C,\F^C)$. Yet it is easy to show that there exists a unique canonical SCM that is equivalent to the PO-SCM, meaning that both SCMs induce identical probabilities of counterfactuals. This result shows that the $PO$ semantics is not incompatible with the Pearlian semantics, but rather refines it, for it can be interpreted as a special case of the Pearlian semantics where one specific canonical SCM is identified as the true underlying model. The justification for this identification is not that we have ruled out all other SCMs based on empirical evidence, but rather that all other SCMs implicitly contain dependence relations that have no meaningful basis in reality and are thus artificial (within the context of causally complete realistic models, to be clear).

\begin{restatable}{proposition}{canon}\label{pro:canon} Given a causal model $M$, there exists a unique canonical SCM $M^{IC}$ such that for any  ${\bf y},{\bf x},{\bf z},{\bf w}$ and ${\bf Y},{\bf X},{\bf Z},{\bf W} \subseteq {\bf V}$
\begin{align*}
P^{IC}({\bf y}_{{\bf x}}, \ldots, {\bf z}_{{\bf w}})=P^{N}({\bf y}_{{\bf x}}, \ldots, {\bf z}_{{\bf w}})
\end{align*}

We say that $M^{IC}$ is the {\em independent canonical} SCM for $M$. 
\end{restatable}

(See the Appendix for proofs of all Theorems.) 

\section{Other Semantics and their Equivalence}\label{sec:equiv}

\subsection{Comparison to Beckers}\label{sec:bec}

In our analysis of the semantics based on the PO-SCM we noted that the independence of the PO variables was an artificial formal tool that we used to compensate for the artificial nature of the counterfactual outcome variables. By doing so we avoid attributing empirical content to artificial modelling devices, as opposed to Pearl's semantics that is based on the response variables, but we still fall short of the principle that all elements of a causal model ought to have a realistic interpretation. Therefore a natural suggestion is to go even further and get rid of all artificial variables altogether, whilst nonetheless maintaining the current semantics. This is precisely what my previous semantics from \cite{beckers25} achieves.\footnote{To be clear, there I did not make explicit that my previous semantics was one for causally complete realistic models only, although I did allude to it by making comments that echo those I made about realism in Section \ref{sec:real}.}

It does so by focussing directly on the relation between the prior and posterior distribution of each $X \in {\bf V}$ conditional on its parents, instead of using the actual outcome variable as a detour. Concretely, prior to the realization of any variables, the conditional distribution given each ${\bf pa_X}$ is $P(X | {\bf pa_X})$. All of these distributions can be interpreted in a forward-looking manner: they express the distribution of $X$ if ${\bf Pa_X}={\bf pa_X}$ is realized. After the realization of some actual ${\bf Pa_X}={\bf pa_X}$, the forward-looking interpretation applies solely to $P(X | {\bf pa_X})$. All distributions $P(X | {\bf pa_X}^*)$ with ${\bf pa_X}^* \neq {\bf pa_X}$ are now to be interpreted as backward-looking, counterfactual, distributions. Once some actual $X=x$ is realized, we obtain the posterior distributions by updating on the evidence that we obtained. The evidence consists of learning that in the {\em actual world}, the causal mechanism that is initiated by ${\bf pa_X}$ is such that it produces $x$. As a consequence, we get the posterior distribution $P(X=x | {\bf pa_X})=1$ for the actually realized ${\bf pa_X}$, whereas the posterior distributions for all {\em counterfactual} worlds remain identical to their prior distribution. 

My previous semantics formalizes this idea by defining the {\em actualized refinement} as an operator on a causal model that updates its probability distribution in the manner just outlined.

\dfn\label{def:ar2} Given a causal model $M$ and some $({\bf u},{\bf v})$ such that $P_M( {\bf u},{\bf v}) >0$, we define the {\em actualized refinement} $M^{({\bf u},{\bf v})}$ as the model  in which $P$ is
replaced by $P^{({\bf u},{\bf v})}$, as follows: for each variable $X \in {\bf V}$ and $(x,{\bf pa_X}) \subseteq ({\bf u},{\bf v})$, the distributions $P(X | {\bf Pa_X})$ are replaced with the distributions $P^{({\bf pa_X},x)}_X(X | {\bf pa_X}')$ that are identical to $P(X | {\bf Pa_X}')$ for all ${\bf pa_X}' \in \R({\bf Pa_X})$ except for ${\bf pa_X}$. Instead, $P^{({\bf pa_X},x)}_X(x' | {\bf pa_X})=1$ if $x'=x$ and $0$ otherwise.
\edfn

Probabilities of basic counterfactuals are then defined as before (Def. \ref{def:pcp}), except that we first replace $P$ with $P^{({\bf u},{\bf v})}$, giving:

\begin{equation*}
 P^B_M({\bf V}={\bf v}^*_{{\bf X} \gets {\bf x}} | {\bf U}={\bf u}, {\bf V}={\bf v})=(P^{({\bf u},{\bf v})})_{do({\bf X} = {\bf x})}( {\bf V}={\bf v}^* | {\bf U}={\bf u})
\end{equation*}

Here $P^B$ is used to denote the current semantics. As before, we write $P^B({\bf y}_{{\bf x}} | {\bf u},{\bf v})$ as shorthand for $P^B_M({\bf Y}={\bf y}_{{\bf X} \gets {\bf x}} | {\bf U}={\bf u}, {\bf V}={\bf v})$ for ${\bf Y},{\bf X} \subseteq {\bf V}$. Probabilities of complex counterfactuals are also defined as before (Def. \ref{def:pc}) -- replacing each $P({\bf a}_{{\bf b}} | {\bf u},{\bf v})$ with $P^B({\bf a}_{{\bf b}} | {\bf u},{\bf v})$ -- and we write these as $P^B({\bf y}_{{\bf x}}, \ldots, {\bf z}_{{\bf w}})$.

Although this semantics is formulated without reference to an underlying SCM and did not rely on using potential outcome variables, when restricted to basic counterfactuals it is equivalent to the novel semantics.

\begin{restatable}{theorem}{basic}\label{thm:basic} Given a causal model $M$, for any  ${\bf y},{\bf x},{\bf v},{\bf u}$ and ${\bf Y},{\bf X} \subseteq {\bf V}$ $$P^B({\bf y}_{{\bf x}} | {\bf u},{\bf v}) = P^N({\bf y}_{{\bf x}} | {\bf u},{\bf v}).$$
\end{restatable}

Importantly, note that ${\bf U}$ here denotes the exogenous variables of $M$, so the expression on the right should not be confused with $P^N({\bf y}_{{\bf x}} | {\bf u}^N,{\bf v})$. As with any SCM, the latter is an extremal probability, whereas $P^N({\bf y}_{{\bf x}} | {\bf u},{\bf v})$ is usually not.

The equivalence between both semantics does not generalize to complex counterfactuals. For a counterexample, we can again take the model from Example \ref{ex:illus}, and consider $P_M((Y=1)_{X=1},(Y=0)_{X=1} | X=0, Y=0)$. I leave it to the reader to verify that $P^B((Y=1)_{X=1},(Y=0)_{X=1} | X=0, Y=0) =p(1-p)$, whereas 
$P^N((Y=1)_{X=1},(Y=0)_{X=1} | X=0, Y=0) = 0$. 

Roughly, the difference between both semantics can be explained as follows. Say we have $P({\bf y}_{{\bf x}} | {\bf u}, {\bf v})=p_1$ and $P({\bf z}_{{\bf w}} | {\bf u}, {\bf v})=p_2$. My previous semantics takes $P({\bf y}_{{\bf x}}, {\bf z}_{{\bf w}}| {\bf u}, {\bf v})=P({\bf y}_{{\bf x}} | {\bf u}, {\bf v})  P({\bf z}_{{\bf w}} | {\bf u}, {\bf v})$ because it interprets such an expression as the statement: given an actual world $({\bf u}, {\bf v})$, if it were the case that ${\bf x}$ then ${\bf y}$ with probability $p_1$ and if it were the case that ${\bf w}$ then ${\bf z}$ with probability $p_2$. The new PO-semantics instead follows the Pearlian semantics in interpreting $P({\bf y}_{{\bf x}}, {\bf z}_{{\bf w}}| {\bf u}, {\bf v})=p$ as the statement: given an actual world $({\bf u}, {\bf v})$, then with probability $p$ if it were the case that ${\bf x}$ then ${\bf y}$ and if it were the case that ${\bf w}$ then ${\bf z}$.  I set aside the discussion of the relative merits of each interpretation for a different occasion, because as it turns out, the equivalence between the two semantics does generalize to all of the common complex counterfactuals that are considered in the literature. For starters, it generalizes to the standard conditional case. 

\begin{restatable}{proposition}{probasic}\label{pro:basic} Given a causal model $M$, for any  ${\bf y},{\bf x},{\bf z}$ and ${\bf Y},{\bf X},{\bf Z} \subseteq {\bf V}$, $P^B({\bf y}_{{\bf x}} | {\bf z}) = P^N({\bf y}_{{\bf x}} | {\bf z})$. 
\end{restatable}

Lastly, the equivalence also holds for the most popular complex counterfactuals appearing in the literature. 

\begin{restatable}{proposition}{pns}\label{pro:pns} Given a causal model $M$, for any binary $X,Y \in {\bf V}$, it holds that $PN^B=PN^N$, $PS^B=PS^N$, and $PNS^B=PNS^N$. (Here PN, PS, and PNS represent \cite{pearl:book2}'s{\em Probability of Necessity}, {\em Probability of Sufficiency}, and {\em Probability of Necessity and Sufficiency}, respectively.)
\end{restatable}

\subsection{Comparison to Galhotra and  Halpern}

\cite{galhotra24}  have recently also proposed a semantics for probabilities of counterfactuals in causal models that take on the form of a Causal Bayesian Network. Furthermore, just as I do, they also include exogenous variables into the graph. Their approach is based on a special type of independence assumption, namely that ``the equations that give the values of a variable for different settings of its parents'' are mutually independent \citep[p. 2]{galhotra24}. Similar to my focus on realistic causally complete models, they argue that their assumption is appropriate in particular when we can observe the exogenous variables, and the model satisfies the causal completeness condition. As I now make clear, their independence assumption is equivalent to the independence of the potential outcome variables that we relied on in Section \ref{sec:poscm}, and as a result their semantics is equivalent to the PO semantics.

Following the standard Pearlian semantics that was presented in Section \ref{sec:pearl}, Galhotra and  Halpern show how the probabilities of counterfactuals in causal models as defined by their semantics can be reduced to their counterparts in SCMs (which can be computed in the standard manner -- Def. \ref{def:pcp}). However, contrary to Pearl, they restrict the class of SCMs to what they call {\em i-compatible} SCMs. They then show that, for a given causal model $M$ and any two i-compatible SCMs $M'$ and $M''$, the probabilities of counterfactuals in $M'$ and $M''$ are identical (see their Theorem 3.7). As a result, it suffices to focus on a single i-compatible SCM that can be efficiently constructed from $M$, and they offer one such construction explicitly. The reader may verify that the first step in their construction is to invoke canonical SCMs (Def. \ref{def:canon}). As a second step, they construct one specific canonical SCM by imposing a special kind of independence property, the result of which looks as follows.

\dfn\label{def:gh}
Given a causal model $M=(\cal S,\G, P)$, we define the {\em $GH$-SCM} $M^{GH}$ as the canonical SCM of $M$ such that for each $X \in {\bf V}$, with parents ${\bf Pa_X} \in \G$, it holds that $$P_{{\bf U}^{C}}(U_X=f_X)=\prod_{\{{\bf pa_X} \in \R({\bf Pa_X})\}} P_M(X=f_X({\bf pa_X}) | {\bf Pa_X} = {\bf pa_X}).$$\edfn

The following result confirms that the $GH$-SCM and the $PO$-SCM are equivalent.

\begin{restatable}{proposition}{gh}\label{pro:gh} Given a causal model $M=(\cal S,\G, P)$, for any  ${\bf y},{\bf x},{\bf z},{\bf w}$ and ${\bf Y},{\bf X},{\bf Z},{\bf W} \subseteq {\bf V}$
\begin{align*}
P^{GH}({\bf y}_{{\bf x}}, \ldots, {\bf z}_{{\bf w}})=P^{N}({\bf y}_{{\bf x}}, \ldots, {\bf z}_{{\bf w}})
\end{align*}
\end{restatable}

Combining these results with those from Sections \ref{sec:poscm} and \ref{sec:bec}, we get:

\cor Given a causal model $M$, for any  ${\bf y},{\bf x},{\bf z},{\bf w},{\bf u}$ with ${\bf Y},{\bf X},{\bf Z},{\bf W} \subseteq {\bf V}$ and ${\bf u} \in \R({\bf U})$:
\begin{itemize}
\item $P^{GH}({\bf y}_{{\bf x}}, \ldots, {\bf z}_{{\bf w}})=P^{N}({\bf y}_{{\bf x}}, \ldots, {\bf z}_{{\bf w}})=P^{IC}({\bf y}_{{\bf x}}, \ldots, {\bf z}_{{\bf w}}).$
\item $P^B({\bf y}_{{\bf x}} | {\bf u},{\bf v}) = P^{GH}({\bf y}_{{\bf x}} | {\bf u},{\bf v})=P^{N}({\bf y}_{{\bf x}} | {\bf u},{\bf v})=P^{IC}({\bf y}_{{\bf x}} | {\bf u},{\bf v})$. 
\item $P^B({\bf y}_{{\bf x}} | {\bf z}) = P^{GH}({\bf y}_{{\bf x}} | {\bf z})=P^{N}({\bf y}_{{\bf x}} | {\bf z})=P^{IC}({\bf y}_{{\bf x}} | {\bf z})$. 
\item for any binary $X,Y \in {\bf V}$, 
\begin{itemize}
\item $PN^B=PN^{GH}=PN^{N}=PN^{IC}$, 
\item $PS^B=PS^{GH}=PS^{N}=PS^{IC}$,
\item $PNS^B=PNS^{GH}=PNS^{N}=PNS^{IC}$.
\end{itemize} 
\end{itemize}
\ecor

\subsection{Comparison to Stochastic Counterfactuals}

\subsubsection{Dawid}

I motivated the current project by relying on \cite{dawid00}'s criticism of the Pearlian semantics, and it is now time to come full circle. After dismissing Pearl's reliance on causal determinism as an unwarranted metaphysical assumption, Dawid briefly considers what a semantics for counterfactuals might look like if we have a causal model that contains all ``sufficient concomitants'' \citep[p. 419]{dawid00}. In our terminology, this is just the assumption that a causal model is complete. Add to that his insistence that we should only use realistic variables -- he speaks of ``genuine'' and ``true concomitants'' -- and we find ourselves in exactly the situation that my semantics is concerned with. Crucially, he then goes on to propose the assumption that ``all variables be treated as conditionally independent across complementary universes, given all the concomitants (which are, of course, constant across universes).'' 

The variables in question are of the form $Y_x({\bf u})$ and $Y_{x'}({\bf u})$, complementarity refers to the fact that $x \neq x'$, the sufficient concomitants are all the exogenous parents ${\bf U}$ of $Y$, and all this within a simplified setting where $X$ and $Y$ are the only two endogenous variables. Thus ${\bf Pa_Y}={\bf U} \cup \{X\}$, so that in our notation Dawid's independence assumption is $P(Y_{{\bf pa_Y}'},Y_{{\bf pa_Y}''})=P(Y_{{\bf pa_Y}'})P(Y_{{\bf pa_Y}''})$ for any ${\bf pa_Y}' \neq {\bf pa_Y}''$, which is the independence assumption of my PO-SCM semantics from Section \ref{sec:poscm}.  \cite{dawid22} return to this proposal, illustrating how it allows for the point-identification of probabilities of causation in the same way as I did in Example \ref{ex:illus2}. Therefore my semantics can be viewed as the formal development of Dawid (and  Musio's) brief suggestion.

\subsubsection{Robins and  Greenland}

Several commentaries to \cite{dawid00}'s wide-ranging criticism of counterfactuals were published together with his article in order to allow the main protagonists in the debate a chance to counter his criticism.  \cite{robins00} take up a defense of counterfactuals as they are understood within the Potential Outcomes framework. Just as does Dawid, they invoke the causal completeness assumption, which they refer to as the ````all causes'' approach'' \citep[p. 432]{robins00}. 

Their analysis of the deterministic case focusses on a causal model with just two binary endogenous variables $X,Y$, and proceeds entirely along the lines of the Pearlian semantics. Concretely, given an equation $Y=f(X,{\bf U})$ for binary variables $X,Y$, they {\em define} the counterfactuals $Y_{X=1}({\bf u}),Y_{X=0}({\bf u})$ as $Y_{x}({\bf u})=f(x,{\bf u})$, and rewrite the equation as $Y=Y_{X}({\bf U})$. This is the same formulation that I used for the PO-SCM, just using slightly different notation.

They then consider a generalization to ``a stochastic counterfactual model'' that satisfies the following properties:

\begin{enumerate}
\item We have ``counterfactual probabilities'' $p_{(X=1, {\bf U})}, p_{(X=0, {\bf U})}$.
\item $P(Y_{x}({\bf u})=1)=p_{(x, {\bf u})}$ if $X=x$ and ${\bf U}={\bf u}$ and undefined otherwise. 
\item $Y=Y_{x}({\bf u})$
\end{enumerate}


As was the case for the PO-SCM, it follows that $P(Y=1 | X=1, {\bf U})=P(Y_{(1,{\bf U})}=1)=p_{(X=1, {\bf U})}$. Finally, when considering a joint distribution over $(Y_{X=1}({\bf U}),Y_{X=0}({\bf U}))$, they make the following independence assumption: $$P(Y_{X=1}({\bf U}),Y_{X=0}({\bf U}) | p_{(X=1, {\bf U})}, p_{(X=0, {\bf U})}) = P(Y_{X=1}({\bf U}) | p_{(X=1, {\bf U})}) P(Y_{X=0}({\bf U}) | p_{(X=0, {\bf U})}).$$ 

Given that, in our definition of a causal model, the probabilities $p_{(X=1, {\bf U})}=P(Y=1 | X=1, {\bf U})$ and $p_{(X=0, {\bf U})}=P(Y=1 | X=0, {\bf U})$ are not variables but fixed features of a model, this assumption is again equivalent to the independence assumption made for the PO-SCM. (To be clear, Robins and  Greenland explicitly point out the similarity to Dawid's proposal as well.)

\subsubsection{Vanderweele and  Robins}

\cite{vanderweele12} discuss stochastic counterfactuals within the context of {\em sufficent cause representations} of a causal model, the deterministic version of which was formally introduced in \cite{vanderweele08} and bears close resemblance to the NESS analysis of causation \cite{wright11,beckers21a}. Concretely they generalize such representations to include the stochastic counterfactuals that were proposed in \cite{robins00}, as just discussed.

As such the details of the sufficient cause representations do not concern us here, what matters for our purposes is simply that a sufficient cause representation consists of decomposing a single potential outcome variable $Y_{\bf Pa_Y}$ into a formula $\phi$ in disjunctive normal form, so that we have the equation $Y=\phi$ instead of $Y=Y_{\bf Pa_Y}$. (Here both $Y$ and all of its endogenous parents are assumed to be binary variables.) Roughly, the idea is that each of the conjunctions appearing in $\phi$ represents a distinct sufficient cause of $Y$, and the entire disjunction is a necessary condition for the occurrence of $Y$. 

Vanderweele and  Robins generalize such representations by adding to each conjunction $C_i$ a random variable $R_i$ that expresses whether or not that particular sufficient cause is in fact sufficient or not, on a given occasion, for a given individual $U$. They consider the assumption that, given an individual $U=u$, such variables $R_i,R_j$ are all mutually independent. As each $R_i$ is associated with a different counterfactual setting, this is entirely in line with my independence assumption. Furthermore, letting ${\bf EPa_Y}$ denote $Y$'s endogenous parents, they note that any structural equation always allows for at least one sufficient cause representation, namely the following: $$Y=\bigvee_{\{{\bf epa_Y} \in \R({\bf EPa_Y})\}} Y_{\bf epa_Y}(U) 1_{\{{\bf EPa_Y} = {\bf epa_Y}\}}.$$ 
Let us call this the {\em canonical sufficient cause representation}. Letting ${\bf XPa_Y}$ denote $Y$'s exogenous parents, and using the notation  $Y_{\bf EPa_Y}({\bf XPa_Y})$ for $Y_{\bf Pa_Y}$, this is just a special case of the representation $Y=Y_{\bf Pa_Y}$ that we used for the PO-SCM (Def. \ref{def:noncanon}). Thus the Potential Outcome SCM is simply a generalization of the canonical sufficient cause representation beyond binary variables and allowing for multiple exogenous parents, combined with the above independence assumption.
 
 \subsubsection{Gelman and  Mikhaeil}

Most recently,  \cite{gelman25} offer a Russian roulette example in which a person has to either spin and then fire a gun with six chambers, one of which contains a bullet ($Z=0$), or do so for a different gun with seven chambers ($Z=1$). The survival probabilities under each option are $1/6$ and $1/7$, and thus we can represent this as $Y=Y_{Z}$ with $P(Y_0)=1/6$ and $P(Y_1)=1/7$. They interpret this model both deterministically, meaning that the probabilities merely reflect our ignorance regarding the determinants of $(Y_{Z=0},Y_{Z=1})$, or stochastically, meaning that the probabilities reflect a random spinning process whose outcome is truly unpredictable. Crucially, they claim that the potential outcomes should be taken to be independent under both interpretations, just as I did for the PO-SCM semantics.

\section{Conclusion}

I have here offered a compromise in the longstanding debate between Pearl and Dawid over the semantics of causal counterfactuals. I did so by taking seriously Dawid's concern that an assumption of universal causal determinism is unwarranted and that causal quantities such as counterfactuals should not depend on unrealistic response variables, and then generalizing the Pearlian semantics accordingly. As a first contribution, I showed that not all realistic and causally complete models are deterministic, and therefore a semantics for irreducibly nondeterministic causal models is indeed needed. I then developed such a semantics by merging the Pearlian framework with the Potential Outcomes framework, and showed it to be equivalent to a previous semantics of mine and another recent proposal. This semantics was also shown to be a formal generalization of the various suggestions for handling stochastic counterfactuals that have appeared throughout the literature more broadly. Although such a wide range of agreement does not by itself prove my semantics to be correct, in the least it suggests that it is based on plausible and widely accepted principles. 

\section*{Appendix}

\canon*

\prf 
For each $X \in {\bf V}$, letting $\R({\bf Pa_X})=\{{\bf pa_X}',\ldots,{\bf pa_X}''\}$, we write $X_{\bf Pa_X}$ for $(X_{{\bf pa_X}'},\ldots,X_{{\bf pa_X}''})$, and similarly for $x_{\bf Pa_X}$. As mentioned above, there is a one-to-one correspondence between the variable $U_X$ and the variables $X_{\bf Pa_X}$, for we can write $U_X=f_X$ as $U_X=x_{\bf Pa_X}$, by reading this as $f_X({\bf pa_X})=x_{\bf pa_X}$ for each ${\bf pa_X} \in \R({\bf Pa_X})$. Similarly, each equation $X = f_X(\bf{Pa_X})$ from the canonical frame $\F^{C}$ corresponds to the equation $X = X_{\bf Pa_X}$ from $\F^{N}$. If we now take $P^{IC}_{{\bf U}^C}(U_X=f_X)=P^N_{{\bf U}^{N}}(X_{\bf Pa_X}=x_{\bf Pa_X})$ for each $X \in {\bf X}$, the result follows directly.
\eprf

\basic* 

\prf It suffices to prove that $P^B({\bf V}={\bf v}^*_{{\bf X} \gets {\bf x}} | {\bf U}={\bf u}, {\bf V}={\bf v})=P^{N}({\bf V}={\bf v}^*_{{\bf X} \gets {\bf x}} | {\bf U}={\bf u}, {\bf V}={\bf v})$. It will be convenient to use the more explicit representation given by Beckers that splits up the former into four cases:
\begin{equation*}
 P^B({\bf V}={\bf v}^*_{{\bf X} \gets {\bf x}} | {\bf U}={\bf u}, {\bf V}={\bf v})=
  \begin{cases}
  0 \text { if }  {\bf x}^* \neq {\bf x}\\
    0 \text { if }  \emptyset \neq \{Y \in {\bf V} \setminus {\bf X} | {\bf epa_Y} = {\bf epa_Y}^* \text{ and } y^* \neq y\}\\
  1\text{ if } \emptyset=\{Y \in {\bf V} \setminus {\bf X} | \bf{epa_Y} \neq \bf{epa_Y}^*\}\\
  \prod_{\{Y \in {\bf V} \setminus {\bf X} | \bf{epa_Y} \neq \bf{epa_Y}^*\}} P_Y(y^* | \bf{epa_Y}^*,\bf{xpa_Y})  \text{ otherwise. }\end{cases}
\end{equation*}
Here for each $Y \in {\bf V}$, ${\bf EPa_Y} = {\bf Pa_Y} \cap {\bf V}$, ${\bf XPa_Y} = {\bf Pa_Y} \cap {\bf U}$, $\bf{epa_Y} \subseteq {\bf v}$, $ \bf{xpa_Y} \subseteq {\bf u}$, $ \bf{epa_Y}^* \subseteq {\bf v}^*$, $y \in {\bf v}$, and $y^* \in {\bf v}^*$.

By the axiom of effectiveness \citep{halpern00}, the RHS evaluates to $0$ whenever ${\bf x} \not \subseteq {\bf v}^*$, and this corresponds to the first case. So we may assume that ${\bf x} \subseteq {\bf v}^*$, and we clarify this assumption by writing  $P^B({\bf V}={\bf v}^*_{{\bf X} \gets {\bf x}^*} | {\bf U}={\bf u}, {\bf V}={\bf v})=P^{N}({\bf V}={\bf v}^*_{{\bf X} \gets {\bf x}^*} | {\bf U}={\bf u}, {\bf V}={\bf v})$. Note also that in both cases this is well-defined only if $P( {\bf u},{\bf v}) >0$. Lastly, as a minor technical point, note that we can assume without loss of generality that each $U \in {\bf U}$ has at least one child. 

Recall that for each $Y \in {\bf V}$ we have the equation $Y = Y_{\bf Pa_Y}$ in $M^N$, and a solution of $M^N$ is a setting of the form $({\bf U}^N={\bf u}^N,{\bf U}={\bf u}, {\bf V}={\bf v}) \in \R({\bf U}^N \cup {\bf U} \cup {\bf V})$. Given some setting $({\bf U}={\bf u}, {\bf V}={\bf v}) \in \R({\bf U} \cup {\bf V})$, for each $Y \in {\bf V}$ there is precisely one $y \in {\bf v}$ and one ${\bf pa_Y} \subseteq ({\bf u},{\bf v})$. Therefore any extension of such a setting into a setting $({\bf U}^N={\bf u}^N,{\bf U}={\bf u}, {\bf V}={\bf v})$ such that for each $Y \in {\bf V}$, $Y_{\bf pa_Y}=y$ and ${\bf Pa_Y}={\bf pa_Y}$, is a solution of $M^N$. Vice versa, given precisely one setting $Y_{\bf pa_Y}=y$ and ${\bf Pa_Y}={\bf pa_Y}$ for each $Y \in {\bf V}$ such that they are all mutually consistent, we can infer precisely one setting $({\bf U}={\bf u}, {\bf V}={\bf v})$ that is consistent with it. (Recall the assumption that each $U \in {\bf U}$ is a parent of some variable.)

Letting ${\bf V}=\{A,\ldots,B\}$, this means that there is a one-to-one correspondence between a setting $({\bf u}, a,\ldots,b)$ and a setting $(a_{\bf pa_A},\ldots,b_{\bf pa_B})$, where ${\bf pa_A} \subset (a,\ldots,b),\ldots,{\bf pa_B}\subset (a,\ldots,b)$. Let ${\bf V} \setminus {\bf X}=\{C,\ldots,D\}$. We likewise get a correspondence between a setting $({\bf u},{\bf v}^*)=({\bf u}, {\bf x}^*,c^*,\ldots,d^*)$ and a setting of the form $({\bf x}^*,c^*_{{\bf pa_C}^*},\ldots,d^*_{{\bf pa_D}^*})$. Using this correspondence for both $({\bf U}={\bf u},{\bf V}={\bf v}^*)$ and $({\bf U}={\bf u}, {\bf V}={\bf v})$, and letting ${\bf X}=\{X^1,\ldots,X^k\}$, we then get

$P^{N}({\bf v}^*_{{\bf X} \gets {\bf x}^*} | {\bf u}, {\bf v}) =P^{N}(({\bf x}^*,c^*,\ldots,d^*)_{{\bf X} \gets {\bf x}^*} | {\bf u}, {\bf v}) =\\
 P^{N}(c^*_{{\bf pa_C}^*},\ldots,d^*_{{\bf pa_D}^*} | x^1_{\bf pa_X^1},\ldots,x^k_{\bf pa_X^k},c_{\bf pa_C},\ldots,d_{\bf pa_D})=\\
\prod_{\{Y \in {\bf V} \setminus {\bf X}\}} P^{N}(y^*_{{\bf pa_Y}^*} | y_{{\bf pa_Y}})=\\
\prod_{\{Y \in {\bf V} \setminus {\bf X} | \bf{pa_Y} = \bf{pa_Y}^*\}} P^{N}(y^*_{{\bf pa_Y}^*} | y_{{\bf pa_Y}})\prod_{\{Y \in {\bf V} \setminus {\bf X} | \bf{pa_Y} \neq \bf{pa_Y}^*\}} P^{N}(y^*_{{\bf pa_Y}^*} | y_{{\bf pa_Y}})=\\
\prod_{\{Y \in {\bf V} \setminus {\bf X} | \bf{pa_Y} = \bf{pa_Y}^*\}} 1_{\{y^*_{{\bf pa_Y}^*}=y_{\bf pa_Y}\}} \prod_{\{Y \in {\bf V} \setminus {\bf X} | \bf{pa_Y} \neq \bf{pa_Y}^*\}} P^{N}(y^*_{{\bf pa_Y}^*})=\\
\prod_{\{Y \in {\bf V} \setminus {\bf X} | \bf{pa_Y} = \bf{pa_Y}^*\}} 1_{\{y^*_{{\bf pa_Y}^*}=y_{\bf pa_Y}\}} \prod_{\{Y \in {\bf V} \setminus {\bf X} | \bf{pa_Y} \neq \bf{pa_Y}^*\}} P(y^* | {{\bf pa_Y}^*})$

{\bf Second Case}: assume that $Z \in {\bf V} \setminus {\bf X}$, $\bf{epa_Z} = \bf{epa_Z}^*$,  and $z^* \neq z$. Since ${\bf xpa_Z} \subseteq {\bf u}$ and ${\bf xpa_Z}^* \subseteq {\bf u}$, we have that  ${\bf xpa_Z} = {\bf xpa_Z}^*$ and thus $\bf{pa_Z} = \bf{pa_Z}^*$. We then have $1_{\{z^*_{{\bf pa_Z}^*}=z_{\bf pa_Z}\}}=1_{\{z^*=z\}}=0$, and the result follows.

{\bf Third Case}: assume that the second case does not hold, and that for each $Y \in {\bf V} \setminus {\bf X}$, $\bf{epa_Y} = \bf{epa_Y}^*$. By the same reasoning as above, this implies that for each $Y \in {\bf V} \setminus {\bf X}$, $\bf{pa_Y} = \bf{pa_Y}^*$, and thus the right product (in the last line above) is empty. By the negation of the second case, we also have that for all $Y \in {\bf V} \setminus {\bf X}$, $y^*=y$, and thus $1_{\{y^*_{{\bf pa_Y}^*}=y_{\bf pa_Y}\}}=1$. From this the result follows.

{\bf Fourth Case}: assume that none of the previous three cases holds. By the negation of the second case, the left product evaluates to $1$. By the negation of the third case, the right product is not empty, and again the result follows.
\eprf

\probasic*

\prf
$P^B({\bf y}_{{\bf x}} | {\bf z}) = P^B({\bf y}_{{\bf x}}, {\bf z}) / P^B({\bf z})=\\
(\sum_{\{{\bf u},{\bf v}\}} P^B({\bf y}_{{\bf x}},{\bf z} | {\bf u},{\bf v}) P( {\bf u},{\bf v}))/P({\bf z})=\\
(\sum_{\{{\bf u},{\bf v}\}} P^B({\bf y}_{{\bf x}} | {\bf u},{\bf v})  P^B({\bf z} | {\bf u},{\bf v}) P( {\bf u},{\bf v}))/P({\bf z})=\\
(\sum_{\{{\bf u},{\bf v} | {\bf z} \subseteq {\bf v}\}} P^B({\bf y}_{{\bf x}} | {\bf u},{\bf v}) P( {\bf u},{\bf v}))/P({\bf z})=\\
(\sum_{\{{\bf u},{\bf v} | {\bf z} \subseteq {\bf v}\}} P^N({\bf y}_{{\bf x}} | {\bf u},{\bf v}) P( {\bf u},{\bf v}))/P({\bf z})=\\
(\sum_{\{{\bf u},{\bf v} | {\bf z} \subseteq {\bf v}\}} P^N({\bf y}_{{\bf x}},{\bf z} | {\bf u},{\bf v}) P( {\bf u},{\bf v}))/P({\bf z})=\\
(\sum_{\{{\bf u},{\bf v}\}} P^N({\bf y}_{{\bf x}},{\bf z} | {\bf u},{\bf v}) P( {\bf u},{\bf v}))/P({\bf z})=\\
P^N({\bf y}_{{\bf x}}, {\bf z}) / P^N({\bf z})=P^N({\bf y}_{{\bf x}} | {\bf z})$
\eprf 

\pns*

\prf The cases for the $PN$ and $PS$ are a direct application of Proposition \ref{pro:basic}. The case for the $PNS$ follows by Lemma 9.2.6 from \cite{pearl:book2}. That this Lemma holds also for the semantics of Beckers follows from the fact that the only causal property that the proof relies on is consistency, and -- as discussed in \citep{beckers25} -- that property holds as well for the semantics of Beckers. 
\eprf

\gh*

\prf Per definition of a canonical SCM, and by Proposition \ref{pro:canon}, it suffices to show that for each $X \in {\bf V}$, $P^{GH}_{{\bf U}^C}(U_X=f_X)=P^{IC}_{{\bf U}^C}(U_X=f_X)$. 

As we did in the proof of Proposition \ref{pro:canon}, we can identify $U_X=f_X$ with $X_{\bf Pa_X}=x_{\bf Pa_X}$, and similarly, $X=f_X({\bf pa_X})$ with $X=x_{\bf pa_X}$. So by Definition \ref{def:gh} it suffices to show that $P^{IC}_{{\bf U}^C}(X_{\bf Pa_X}=x_{\bf Pa_X}) = \prod_{\{{\bf pa_X} \in \R({\bf Pa_X})\}} P_M(X = x_{\bf pa_X} | {\bf Pa_X} = {\bf pa_X})$. 

$P^{IC}_{{\bf U}^C}(X_{\bf Pa_X}=x_{\bf Pa_X}) \stackrel{\text{ (1) }}{=} P^N_{{\bf U}^{N}}(X_{\bf Pa_X}=x_{\bf Pa_X})\stackrel{\text{ (2) }}{=}
 \prod_{\{{\bf pa_X} \in \R({\bf Pa_X})\}}  P^{N}_{{\bf U}^{N}}(X_{\bf pa_X}=x_{\bf pa_X})\stackrel{\text{ (3) }}{=}
 \prod_{\{{\bf pa_X} \in \R({\bf Pa_X})\}}  P_M(X=x_{\bf pa_X}) | {\bf Pa_X}={\bf pa_X}).$

Here (1) follows from the proof of Proposition \ref{pro:canon}, (2) follows from the independence of exogenous variables in an SCM, and (3) follows from Definition \ref{def:noncanon}.
\eprf

\bibliographystyle{spbasic} 
\bibliography{allpapers}

\end{document}

%% file: defn.tex
\newcommand{\thm}{\begin{theorem}}
\newcommand{\lem}{\begin{lemma}}
\newcommand{\pro}{\begin{proposition}}
\newcommand{\dfn}{\begin{definition}}
\newcommand{\rem}{\begin{remark}}
\newcommand{\xam}{\begin{example}}
\newcommand{\cor}{\begin{corollary}}
\newcommand{\prf}{\noindent{\bf Proof:} }
\newcommand{\ethm}{\end{theorem}}
\newcommand{\elem}{\end{lemma}}
\newcommand{\epro}{\end{proposition}}
\newcommand{\edfn}{\bbox\end{definition}}
\newcommand{\erem}{\bbox\end{remark}}
\newcommand{\exam}{\bbox\end{example}}
\newcommand{\ecor}{\end{corollary}}
\newcommand{\eprf}{\bbox\vspace{0.1in}}
\newcommand{\beqn}{\begin{equation}}
\newcommand{\eeqn}{\end{equation}}

\newcommand{\bbox}{\vrule height7pt width4pt depth1pt}

\newcommand{\clm}{\begin{claim}}
\newcommand{\eclm}{\end{claim}}

\newcommand{\sat}{\models}

\newcommand{\union}{\cup}

\renewcommand{\phi}{\varphi}

\newcommand{\F}{{\cal F}}
\newcommand{\G}{{\cal G}}

\newcommand{\R}{{\cal R}}

\newcommand{\ol}{\setlength{\itemsep}{0pt}\begin{enumerate}}
\newcommand{\eol}{\end{enumerate}\setlength{\itemsep}{-\parsep}}
\newcommand{\ul}{\setlength{\itemsep}{0pt}\begin{itemize}}
\newcommand{\dl}{\setlength{\itemsep}{0pt}\begin{description}}
\newcommand{\edl}{\end{description}\setlength{\itemsep}{-\parsep}}
\newcommand{\eul}{\end{itemize}\setlength{\itemsep}{-\parsep}}